\documentclass[letterpaper, 10 pt, conference]{ieeeconf}  

\IEEEoverridecommandlockouts                              

\usepackage{graphics} 
\usepackage{epsfig} 
\usepackage{subfig}
\usepackage{marvosym}
\usepackage{hyperref}
\usepackage{amsmath} 

\title{\LARGE \bf

Web-based Augmented Reality with Auto-Scaling and Real-Time Head Tracking towards Markerless Neurointerventional Preoperative Planning and Training of Head-mounted Robotic Needle Insertion

% ...
}

\author{Hon Lung Ho$^{1}$*, Yupeng Wang$^{1}$*, An Wang$^{1}$*, Long Bai$^{1}$, 
Hongliang Ren$^{1}$\textsuperscript{\Letter},~\IEEEmembership{Senior Member,~IEEE}
\thanks{This work was supported in part by the Hong Kong Research Grants Council (RGC)-General Research Fund (GRF) 14211420, RGC - Collaborative Research Fund under grant CRF C4026-21GF, in part by the Hong Kong Research Grants Council (RGC) Research Impact Fund under grant R4020-22, and in part by the Guangdong Basic and Applied Basic Research Foundation (GBABF) under Grant 2021B1515120035. (\textit{Corresponding author: Hongliang Ren, hlren@ieee.org.})}
\thanks{* H. Ho, Y. Wang, and A. Wang are the co-first authors.}
\thanks{$^{1}$ H. Ho, Y. Wang, A. Wang, L. Bai, and H. Ren are with the Department of Electronics Engineering, The Chinese University of Hong Kong, Hong Kong SAR, China; and CUHK Shenzhen Research Institute, Shenzhen, China.
       }%
}

\begin{document}

\maketitle
\thispagestyle{empty}
\pagestyle{empty}

\begin{abstract}

Neurosurgery requires exceptional precision and comprehensive preoperative planning to ensure optimal patient outcomes. Despite technological advancements, there remains a need for intuitive, accessible tools to enhance surgical preparation and medical education in this field. Traditional methods often lack the immersive experience necessary for surgeons to visualize complex procedures and critical neurovascular structures, while existing advanced solutions may be cost-prohibitive or require specialized hardware.
This research presents a novel markerless web-based augmented reality (AR) application designed to address these challenges in neurointerventional preoperative planning and education. Utilizing MediaPipe for precise facial localization and segmentation, and React Three Fiber for immersive 3D visualization, the application offers an intuitive platform for complex preoperative procedures. A virtual 2-RPS parallel positioner or ``Skull-Bot" model is projected onto the user's face in real-time, simulating surgical tool control with high precision.
Key features include the ability to import and auto-scale head anatomy to the user's dimensions and real-time auto-tracking of head movements once aligned. The web-based nature enables simultaneous access by multiple users, facilitating collaboration during surgeries and allowing medical students to observe live procedures. A pilot study involving three participants evaluated the application's auto-scaling and auto-tracking capabilities through various head rotation exercises.
This research contributes to the field by offering a cost-effective, accessible, and collaborative tool for improving neurosurgical planning and education, potentially leading to better surgical outcomes and more comprehensive training for medical professionals. The source code of our application is publicly available at~\url{https://github.com/Hillllllllton/skullbot_web_ar}.

\end{abstract}

\section{INTRODUCTION}



Minimally invasive neurosurgery (MIN) procedures encompass sophisticated techniques that entail precise manipulation of surgical tools within the delicate brain tissue through minimal cranial incisions \cite{b1}. Any slight misplacement of these tools during the procedure can lead to severe bleeding and significant neurological complications, endangering the patient’s health \cite{b2}. Consequently, achieving meticulous localization of neural structures is crucial, typically based on computed tomography (CT) or magnetic resonance imaging (MRI) for guidance to specific target sites \cite{b1}. However, the direct correlation between CT/MRI imaging and the unique anatomy of the patient's brain can be non-intuitive \cite{b3}, necessitating the implementation of a sophisticated support system \cite{b4} \cite{b5} for precise localization and registration during surgery. 

Currently, most preoperative planning and medical education still rely on static 2D images and textual descriptions, lacking the sense of immersion and interactivity~\cite{b5,b9}. By providing intuitive visualization, Augmented Reality (AR) can highlight critical anatomical structures or pathologies that are hidden or challenging to find and finally guide doctors to perform surgeries more accurately \cite{b8}. AR, which can directly overlay the brain anatomy over the patient's head, holds the potential to revolutionize the pre-surgical planning and educational frameworks within neurosurgery~\cite{b7,b10}. While traditional methods have limitations in terms of time and location, AR technology in healthcare can improve safety, efficiency, doctor-patient relationships, and the invention of new surgical methods, benefiting patient outcomes and fostering innovation \cite{b6}. Therefore, the AR visualization can play a significant role in the preoperative planning of the neurosurgery.

There exist two primary types of AR applications: one relies on head-mounted devices (HMD) for observation, while the other permits direct viewing on mobile apps or websites. While hardware-based AR implementation is recognized for its costliness and lack of flexibility, web-based AR negates additional hardware requirements, facilitating effortless access to AR content on personal smartphones or tablets~\cite{b14}. Moreover, when it comes to surgery, multiple users can access AR content through a single web URL, creating a collaborative platform that enhances the cooperation among surgeons \cite{b16}. Furthermore, certain medical students can actively engage in surgeries without disrupting the operating surgeon \cite{b19}, thereby significantly improving medical learning opportunities. Therefore, we focus on developing and deploying a web AR application for preoperative planning and education of neurosurgery in this paper.

\begin{figure*}
    \centering
    \includegraphics[width=.95\linewidth]{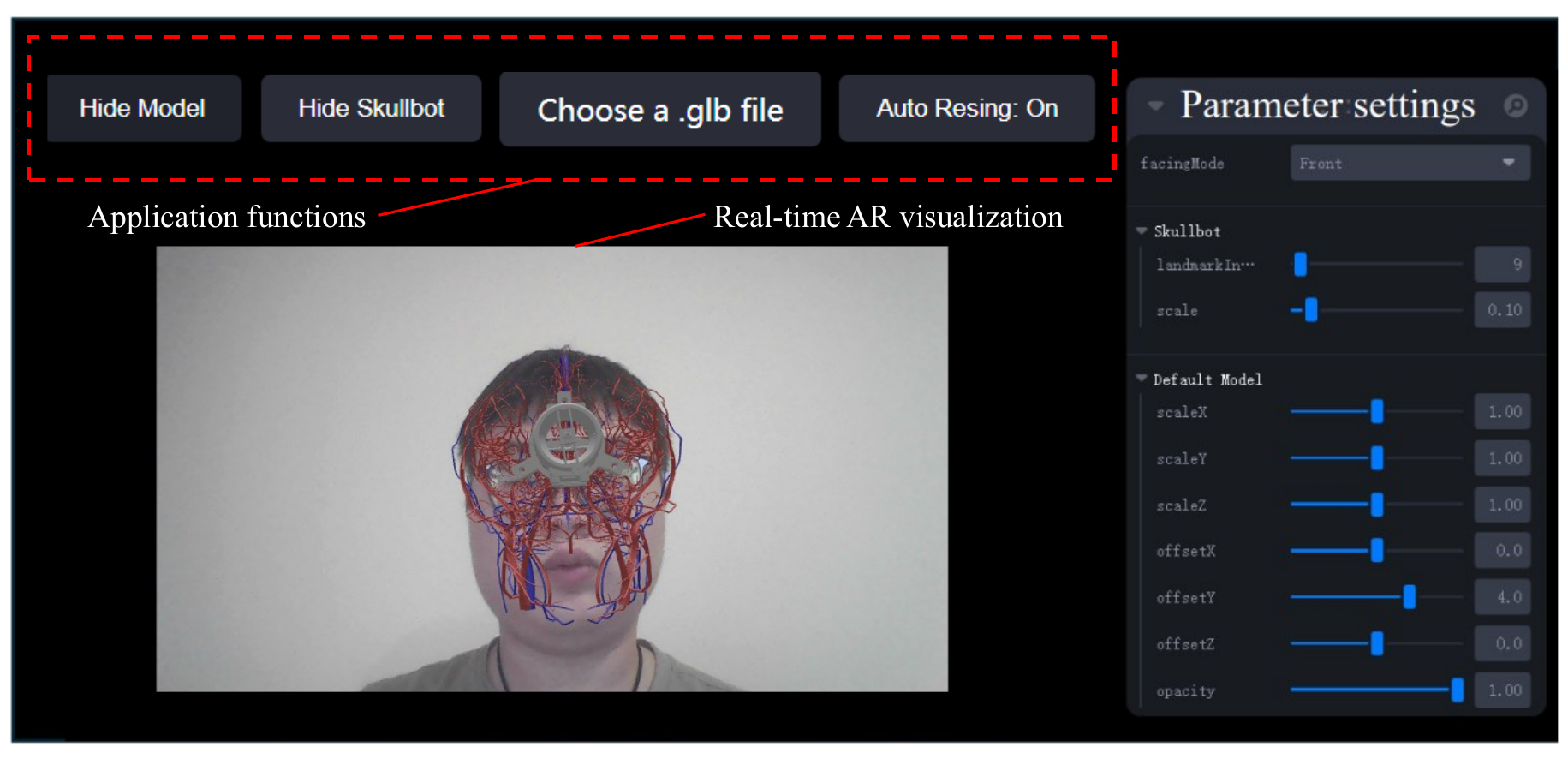}
    \caption{Layout of the web-based AR application.
    The interface includes three distinct regions, each serving a specific purpose. The upper region is the function block with several buttons, which controls the import/hide of the 3D model and enables/disables auto-scaling. The middle region is a window for displaying the AR content. The right region is a panel for parameter settings, which can be used to manually modify the imported model.}
    \label{fig:layout}
\end{figure*}


The main objective of our AR application is to accurately superimpose virtual models onto users' heads. Leveraging Google's MediaPipe \cite{b6}, a robust and widely-used framework for face landmark detection and segmentation, developers can efficiently manage resources while maintaining high-quality results \cite{b15}.
Renowned for its precision and utility in medical applications, MediaPipe has been extensively validated through research, demonstrating 95\% consistency limits and comparable accuracy to traditional measurement tools such as the universal goniometer and digital angle ruler \cite{b11}. Integrating artificial intelligence (AI) in AR has further streamlined automated tracking, significantly reducing preparation time for surgical procedures \cite{b20}.


In minimally invasive neurosurgery, two critical factors are essential: precise visualization of target brain structures and accurate control over surgical instruments. Integrating robotics in surgery has significantly enhanced operating efficiency and safety by reducing operating times, minimizing blood loss, and accelerating patient recovery \cite{b18} \cite{b17}. The 2-RPS parallel positioner (Skull-Bot) is a notable head-mounted robot for needle insertion, achieving remarkable open-loop positioning accuracy within ±1 mm in neurosurgical procedures \cite{b11} \cite{b12}. By combining the Skull-Bot with web-based augmented reality visualization, we aim to further refine the preoperative process with improved intuitiveness.

As shown in Fig.~\ref{fig:layout}, this paper introduces a novel marker-free web-based AR application that can assist neurosurgical preoperative planning and training such as head-mounted robotic needle insertion procedures. 
By leveraging facial feature detection and immersive 3D visualization, our system enables users to import and overlay 3D head anatomy structures in real-time, simulating precise surgical tool placement. This innovative approach aims to streamline pre-surgical training and planning, ultimately reducing the time required for surgeons and medical students to become proficient with specific tools. The contributions of this research are threefold:
\begin{itemize}
\item Development of a markerless web-based AR platform: Our system integrates facial feature detection and achieves immersive 3D visualization, providing an intuitive interface for neurosurgical preoperative planning and education.
\item Auto-scaling and tracking capabilities: The application features auto-rescaling to accommodate different head dimensions and enables fast and precise auto-tracking of head movements, enhancing the overall user experience.
\item Evaluative validation: A qualitative and quantitative assessment involving three users demonstrates the high performance and quality of our system with over 80\% overlap IoU, showcasing its potential to improve neurosurgery procedures.
\end{itemize}


\section{METHODOLOGY}

\begin{figure}
    \centering
    \includegraphics[width=\linewidth]{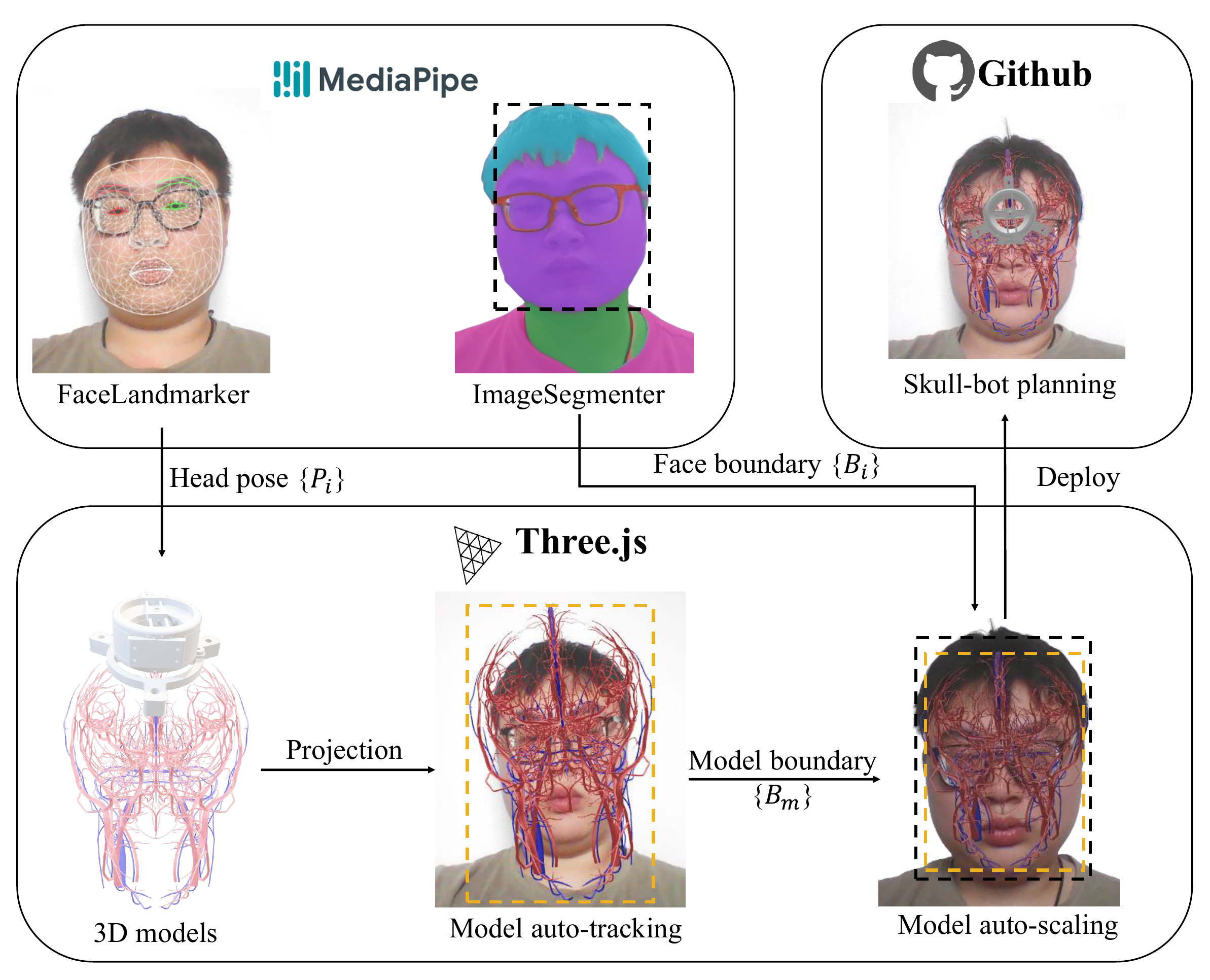}
    \caption{Overall working mechanism of our AR application.}
    \label{fig:workflow}
\end{figure}

\subsection{Application Overview}
The application was developed using Three.js \cite{b13}, a JavaScript library that enables the creation and rendering of 3D objects in web browsers via WebGL. This powerful tool simplifies the development of complex visual effects, allowing for the seamless integration of 3D elements into web-based interfaces. With Three.js, developers can easily create interactive 3D experiences by scaling, rotating, and animating 3D objects directly within the browser. 
Furthermore, React Three Fiber\footnote{\url{https://github.com/pmndrs/react-three-fiber}} (R3F) was employed as a React renderer for Three.js~\cite{b13}. R3F provides an intuitive interface to work with Three.js, seamlessly integrating it into the React ecosystem. This integration not only streamlines the development process but also leverages React's state management capabilities, significantly reducing development time.


The development environment was established by integrating two key libraries: React Three Fiber (R3F) for 3D scene management and MediaPipe \cite{b6} for facial landmark detection. A webcam feed was positioned on a distant plane using R3F, while a 3D model was placed closely in front of the perspective camera offered by Three.js. This setup effectively simulates an augmented reality (AR) environment, rendering the 3D model between the viewer and the webcam plane. The resulting AR simulation enables users to interact with virtual objects as if they were real-world objects, enhancing the overall user experience.
The developed application can be deployed locally or hosted by a server. We have deployed it on GitHub\footnote{\url{https://hillllllllton.github.io/fyp_AR/}} to provide users with convenient access to the AR application from anywhere with an internet connection and a webcam. This deployment strategy not only
facilitates user engagement but also promotes accessibility, allowing users to explore the AR environment at their own pace.

The AR application consists of three main components, as illustrated in Fig.~\ref{fig:layout}. Firstly, an intuitive interface is provided through a set of buttons that enable users to access various functions within the application. This includes the ability to import different 3D head models, which can be customized to display various anatomy structures for educational purposes. Additionally, the skull-bot model can be visualized, facilitating planning and preparation for surgical operations.
The imported 3D model can also be adjusted using the parameters panel, built utilizing Leva\footnote{\url{https://github.com/pmndrs/leva}}, a powerful React widget library that streamlines parameterization and configuration. This feature allows users to fine-tune the model's appearance and behavior, ensuring a precise representation of the anatomy. Finally, the camera view is seamlessly overlaid with the 3D model in real-time, providing an immersive augmented reality experience.

\subsection{AR Application Implementation}
Accurate pre-operative planning requires a precise overlay of the virtual 3D model onto the user's head. This necessitates the implementation of a robust auto-tracking system, which continuously monitors the user's head
position to ensure seamless alignment with the model.
Moreover, the size of the imported 3D virtual model may not match the user's head dimensions, requiring an adaptive solution that adjusts the model's scale accordingly. To address this challenge, an auto-scaling mechanism was developed, which scales the model in real-time based on the user's head size. The successful implementation of these two core functions is critical for a seamless user experience. The general workflow is visualized in Fig.~\ref{fig:workflow}, and we will elaborate on it in detail in the following sections.



\subsubsection{Head auto-tracking}
The FaceLandmarker model from MediaPipe processes webcam footage to detect facial landmarks in real-time. These landmarks serve as a guide for positioning and orienting the 3D model within the AR environment, enabling precise registration between the virtual model and the user's face. 
To ensure seamless tracking of the face, we employ the face mesh, a 4x4 matrix $P_i$, to dynamically adjust the 3D model's orientation and position in real-time. This sophisticated approach ensures that the model accurately overlays and moves with the user's face, providing an immersive AR experience. 

Based on the pose $P_i$, as well as the imported mesh, the surface points $P_w$ can be determined. The application projects the surface points onto a 2D plane, allowing users to visualize it within the web interface. To achieve this projection, we assume that the 2D plane is located at $z_c$ ahead of the camera. By leveraging the properties of perspective projection, we can derive the following equation:

\begin{equation}
    z_c p_c = K T P_w,
\end{equation}
where $p_c$ represents the 2D location of the point in pixels, $K$ is the camera intrinsics matrix that encodes information about the camera's focal length and image sensor dimensions, $T$ is the transformation from the camera frame to the world frame, and $P_w = [X, Y, Z]^T$ denotes the 3D position of the model in the world coordinate system.

Given that we assume the camera frame is equivalent to the world frame, the transformation matrix $T$ becomes an identity matrix, and the depth value $z_c$ reduces to the 3D coordinate $Z$. Furthermore, we utilize the default camera intrinsics provided by three.js. Under these conditions, the equation simplifies to:
\begin{equation}
    p_c = K P_w / z_c = K P_n,
\end{equation}
where the normalized 3D position $P_n$ can be defined as:
\begin{equation}
    P_n = \left[
 \begin{array}{c}
     X / Z \\
     Y / Z \\
     1
 \end{array}
 \right].
\end{equation}
This normalized form can be used to complete the projection and achieve accurate 2D rendering.

\subsubsection{Model auto-rescaling}
When the 3D model is projected onto the 2D image, it is bounded by a rectangular $B_m$. However, its width ($w_m$) and height ($h_m$) may not accurately fit the user's head. To address this issue, we can re-scale the virtual model to achieve a better match between the virtual and real worlds.

The ImageSegmenter from MediaPipe is used to segment the user's head in real-time, providing another boundary $B_i$. The width ($w_i$) and height ($h_i$) of this segmentation are more accurately tailored to fit the user's head. To scale the 3D model boundaries to match these dimensions, we calculate the scale factors:
\begin{align}
    s_w &= w_i / w_m, \\
    s_h &= h_i / h_m.
\end{align}
With these scale factors, the projection equation becomes:
\begin{equation}
    p_c = K S P_n,
\end{equation}
where $S$ is a diagonal scale matrix defined as:
\begin{center}
    $S = \begin{bmatrix}
        s_w & 0 & 0 \\
        0 & s_h & 0 \\
        0 & 0 & 1
    \end{bmatrix}$.
\end{center}
This allows us to auto-scale the 3D model based on head size, achieving a precise virtual-to-real matching for different users.

\subsubsection{Skull-Bot visualization}
In addition to the previously mentioned features, our system also utilizes the skull and the 2-RPS parallel positioner (Skull-Bot) \cite{b11}. The Skull-Bot is particularly well-suited for applications requiring exact placement or
manipulation within constrained spaces, making it an ideal choice for medical or surgical procedures where precision is crucial.
Given its precise positioning requirements, setting up the Skull-Bot can be a complex task. However, our system allows users to simulate the placement of the Skull-Bot for pre-planning surgery and educational training purposes before actually using the machine. This capability enables users to practice and refine their skills in a virtual environment, reducing the risk of errors and improving overall efficiency.



\subsection{Additional Functions}
In addition to its core capabilities, our system includes several additional features designed to enhance the user experience.

\subsubsection{Model visibility toggle}

To further empower users, we have implemented a feature that allows them to toggle the visibility of 3D models on demand. By incorporating a user-friendly button into the interface, users can easily hide or show models as needed, making the application adaptable to different learning and exploration scenarios. This simple yet effective functionality provides users with greater control over their interaction with the application, enabling them to focus on specific aspects of an augmented reality (AR) environment or simplify complex scenes for clearer understanding and guidance.

\subsubsection{Camera selection capabilities}
This application provides a flexible camera selection feature that allows users to seamlessly switch between the front and rear cameras on their device, whether it is a tablet or mobile phone. This capability is particularly valuable in educational and presentation settings, where users often need to toggle perspectives quickly.

\subsubsection{Manual model customization}
In addition to auto-rescaling, our application offers a high degree of manual control over the 3D model, allowing users to personalize their experience and achieve precise alignment. With our intuitive tools, users can manually scale, position, and orient the model in three-dimensional space, ensuring that it perfectly matches their face. Furthermore, users can adjust the model's transparency to see through it to the underlying physical structure. This feature is particularly useful in educational environments or real-world surgery, where visualizing underlying anatomy is crucial.


\subsubsection{Custom model import}

To expand its versatility and cater to diverse educational and practical applications, our software allows users to import their own GLB (.glb) models. This feature empowers users to create highly customizable AR experiences that can accommodate specialized or proprietary models.
With this capability, users can seamlessly integrate their own models into the existing interface, giving them access to the same level of control as preloaded models. Specifically, they can position, scale, and adjust the transparency of their imported model with ease. 
This feature is particularly beneficial for professionals in the medical field, who can create personalized simulations and educational models tailored to their specific needs.

\begin{figure*}
    \centering
    \includegraphics[width=.85\linewidth]{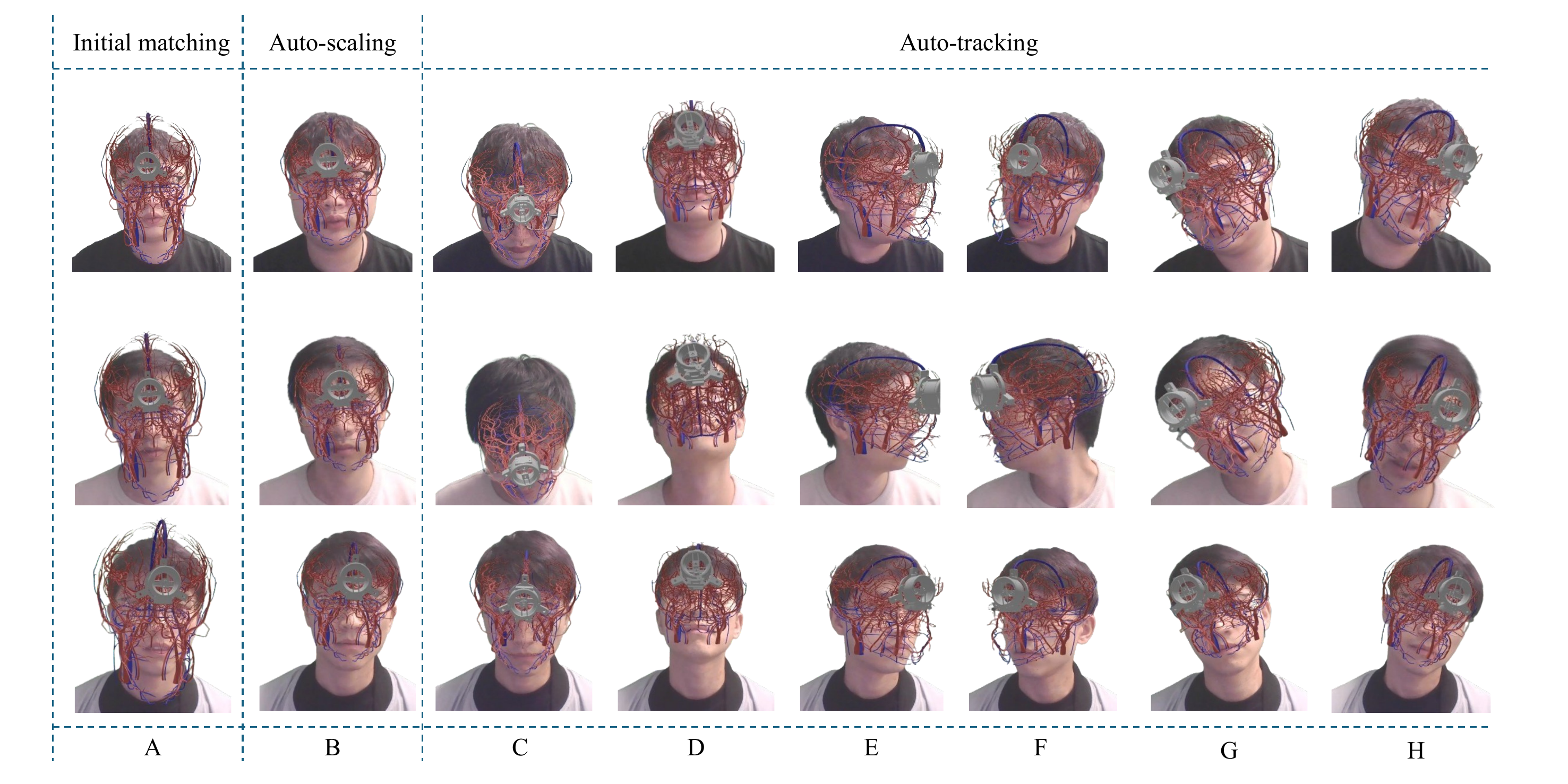}
    \caption{Visualization of the scaling and tracking results. The virtual model can auto-scale itself (A to B) and track head movements automatically. The participants rotated their heads in the direction of Pitch (C, D), Yaw (E, F), and Roll (G, H).}
    \label{fig:quali_res}
\end{figure*}

\begin{figure*}
    \centering
    \includegraphics[width=.85\linewidth]{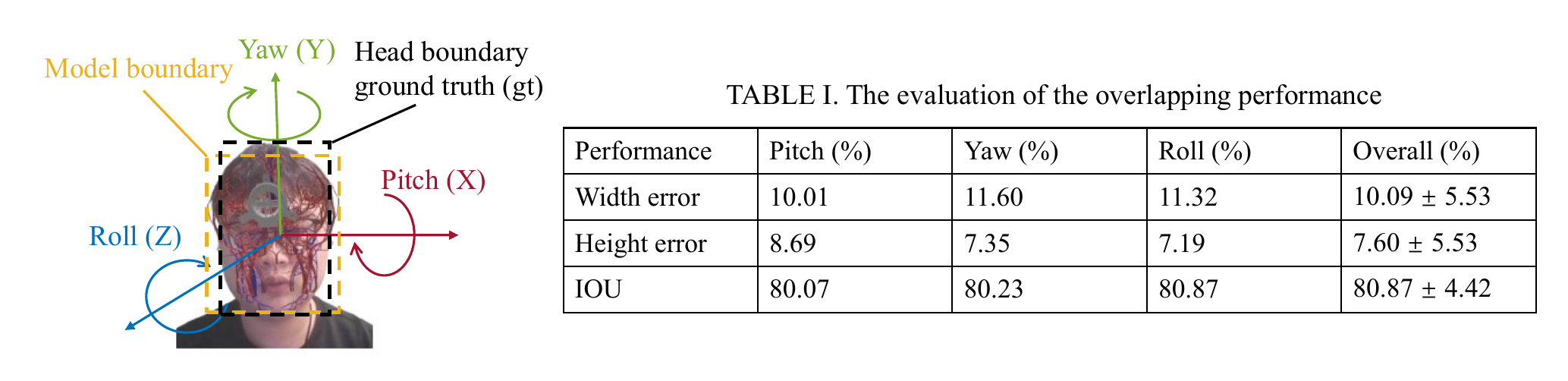}
    \caption{Experimental quantification results. The user's head is rotated in 3 DOF to roughly 45 degrees. For each DOF, the overlap error of width, height, and IOU are calculated. The application generally achieves decent tracking performance.}
    \label{fig:quant_res}
\end{figure*}

\section{EXPERIMENTS AND RESULTS}
\subsection{Experiment Setup}
The AR application was deployed on a computer with an NVIDIA RTX3050 GPU, utilizing 0.4GB of shared memory resources. A detailed 3D model of blood vessels in the head was imported into the application as the test subject.
To evaluate its performance in real-world scenarios, three participants with varying head sizes were recruited for the experiment. 
The initial setup involved overlapping the virtual model on the participant's head and scaling it to match their individual head size. Subsequently, each participant rotated their head around its axis in pitch, yaw, and roll direction by roughly 45 degrees, as demonstrated in the left of Fig.~\ref{fig:quant_res}. Finally, to assess the application's tracking error, we drew the boundaries of both the model ($B_m$) and the participant's head ($B_i$).

\subsection{Evaluation Metrics}
The experiment employed two key metrics to evaluate the AR application's performance: positional accuracy and Intersection over Union (IOU). The former is calculated as the difference between the width and height of the participant's head and the projected 3D model. IOU measures the overlap performance by comparing the intersection area of the model boundary ($B_m$) with the participant's head boundary ($B_i$). A higher IOU value indicates better overlapping accuracy.
In this experiment, we used the participant's head boundary ($B_i$) as the ground truth (gt) for evaluation purposes.

\subsubsection{Width Error And Height Error}
To assess the accuracy of the model scaling, we evaluated its performance in two dimensions: width and height. The scale factor was applied to match the initial model size with that of the user's head.
We defined the error in width ($E_w$) and height ($E_h$) direction as follows:
\begin{align}
    E_w &= w_m / w_i \times 100\%, \\
    E_h &= h_m / h_i \times 100\%,
\end{align}
where $w_i$ and $h_i$ are the width and height of the user's head boundary ($B_i$), respectively, and $w_m$ and $h_m$ are the width and height of the model boundary ($B_m$), respectively.

\subsubsection{Intersection over Union}
In addition to evaluating the scaling accuracy, we also calculated the IOU metric to assess the application's tracking ability. The IOU is defined as:
\begin{equation}
    IOU = {{B_i \cap B_m} \over {B_i \cup B_m}}
\end{equation}
The IOU value represents the ratio of the intersection area to the union area of the two boundaries.



\subsection{Results And Analysis}
The experiment revealed promising results, as demonstrated in Fig.~\ref{fig:quali_res}.
Upon startup, the application would first make an initial match between the head and the virtual model. The virtual model was then scaled to the correct size based on the segmentation result, as demonstrated from columns A to B in Fig.~\ref{fig:quali_res}.  The scaled model effectively tracked the head rotations, achieving satisfactory outcomes despite variations in face shape, hairstyle, and head size among participants. 

The analysis of error metrics revealed moderate disparities between the scaled model and actual head measurements. Specifically, as shown in Table I, the width error was reported as 10.09 ± 5.53\% (mean ± standard deviation), while the height error was 7.60 ± 5.53\%. 
Rotations in the yaw direction had the most significant effect on width error, suggesting that this axis is particularly challenging for accurate model scaling. Conversely, rotations in the pitch direction exerted a more pronounced influence on height error.

Moreover, the IOU values as shown in Table I revealed a relatively high level of agreement between the ground truth and model boundary across different dimensions. The average IOU value of 80.87 ± 4.42\% suggests that the application effectively overlays the virtual model onto the user's head. Notably, rotation in various directions has a similar impact on the IOU metric, indicating robustness against different head orientations.

\section{DISCUSSION AND CONCLUSION}

The development and deployment of our AR application for neurointerventional preoperative planning mark a significant step forward in the integration of medical technology and education. By harnessing the power of MediaPipe~\cite{b6} for facial landmark detection and segmentation, our immersive marker-free 3D visualization approach offers a novel and engaging platform for training and preparation in neurosurgery procedures, i.e., head-mounted robotic needle insertion with Skull-Bot. The current emphasis of this paper is on implementing the registration of a virtual head model in web-based AR. In the future, the real Skull-Bot can be positioned according to AR planning, showcasing its feasibility in actual surgical scenarios.

The experiments conducted on diverse participants demonstrate that the application performs reasonably well in auto-scale accuracy, with IOU scores indicating a strong degree of overlap between the ground truth and model boundary. While these results are promising, further refinements to the scaling algorithm, optimization of the boundary detection method, and fine-tuning of localization accuracy are crucial for continued improvement.

Notably, our application's reliance on facial tracking provided by MediaPipe \cite{b6} poses limitations in certain situations, such as hind-brain surgery training where the face may not be visible or accessible. To broaden its utility and versatility across various surgical procedures, future developments should focus on reducing this dependency and exploring alternative tracking methods or technologies.

In conclusion, the successful integration of auto-scale and auto-tracking capabilities into our markerless web-based AR application has significant implications for preoperative planning, enhancing the situation awareness of surgeons and medical professionals. By addressing the aforementioned limitations and building upon existing successes, we envision a more comprehensive and accessible platform for neurosurgery preoperative planning, ultimately enhancing patient care and outcomes.



\addtolength{\textheight}{-12cm}   





\section*{ACKNOWLEDGMENT}
We would like to express our gratitude to Dr. Lalithkumar Seenivasan for his invaluable discussions and to Ruijie Tang for participating in the experiments.



\bibliographystyle{IEEEtran}
\bibliography{mybib}

\end{document}